# Strong Faithfulness and Uniform Consistency in Causal Inference

Jiji Zhang
Philosophy Department
Carnegie Mellon University
Pittsburgh, PA 15213
jiji@andrew.cmu.edu

Peter Spirtes
Philosophy Department
Carnegie Mellon University
Institute for Human & Machine Cognition
University of West Florida
ps7z@andrew.cmu.edu

## Abstract

A fundamental question in causal inference is whether it is possible to reliably infer manipulation effects from observational data. There are a variety of senses of asymptotic reliability in the statistical literature, among which the most commonly discussed frequentist notions are pointwise consistency and uniform consistency (see, e.g. Bickel, Doksum [2001]). Uniform consistency is in general preferred to pointwise consistency because the former allows us to control the worst case error bounds with a finite sample size. In the sense of pointwise consistency, several reliable causal inference algorithms have been constructed under the Markov and Faithfulness assumptions [Pearl 2000, Spirtes et al. 2001]. In the sense of uniform consistency, however, reliable causal inference is impossible under the two assumptions when time order is unknown and/or latent confounders are present [Robins et al. 2000]. In this paper we present two natural generalizations of the Faithfulness assumption in the context of structural equation models, under which we show that the typical algorithms in the literature (in some cases with modifications) are uniformly consistent even when the time order is unknown. We also discuss the situation where latent confounders may be present and the sense in which the Faithfulness assumption is a limiting case of the stronger assumptions.

## 1 INTRODUCTION

### 1.1 CAUSAL INFERENCE

We consider the kind of causal inference in the literature that predicts the effects of manipulations (or the "do" operator in Pearl [2000]) from non-experimental data [Spirtes et al. 2001]. Such inference typically involves two steps: discovery of causal structures, represented by directed acyclic graphs (DAGs)[1], and identification of causal parameters. There are two main approaches in the causal discovery step: constraint-based approach and Bayesian approach, of which we focus on the former as we are going to discuss the frequentist notions of consistency. The basic idea of the constraint-based approach is to test the conditional independence relations among the observed variables, which, under certain assumptions, put some graphical constraints on the possible causal structures. The two commonly adopted assumptions are the Markov and Faithfulness assumptions. The **Markov assumption** says that every variable is independent of its non-effects conditional on its direct causes, which is just the (local) Markov property of DAGs. The **Faithfulness assumption** says that no conditional independence relations other than the ones entailed by the Markov assumption are present in the population distribution. Since the conditional independence relations entailed by the Markov assumption correspond exactly to d-separation [Pearl 1988], these two assumptions together translate the conditional independence relations in the population distribution to d-separation constraints on the possible causal graphs.

The output of the constraint-based algorithms is thus a *set* of causal graphs compatible with background knowledge that share the same d-separation structures (or say, entail the same conditional independence relations) among the observed variables, which is usually called a Markov equivalence class. A causal quantity is identifiable with respect to a set of causal graphs if given any graph in the set, the causal quantity can be written uniquely in terms of some identifiable statistical quantities. Once a causal parameter is identified, inference concerning that parameter is just ordinary statistical inference.

---

[1] In general directed (cyclic) graphs can be used to represent causal systems that might have feedback. There are (constraint-based) algorithms of discovering causal graphs with (possibly) cycles assuming no latent confounders [Richardson 1996], of which the results in this paper still hold.

In this paper we confine the discussion to one of the most commonly used parameterizations of causal graphs: linear structural equation models (LSEMs), in which the structural coefficients can be easily interpreted as the direct manipulation effects. Specifically, given a DAG, each arrow is assigned a coefficient so that every variable can be written as a linear function of its parents plus a Gaussian error[2]. Usually the variables are standardized for the sake of interpreting the structural coefficients. Without loss of generality, we consider standardized LSEMs in what follows.

Finally, there is an important assumption called *causal sufficiency* that can dramatically simplify causal discovery. A causal system (i.e. a set of observed variables) is causally sufficient if no common cause of any two variables in the system is left out. In more plain words, causal sufficiency assumes that there are no latent confounders of any two observed variables. For simplicity we present our main results under this assumption, but it is not essential as we point out later.

### 1.2 CONSISTENCY OF TESTS

Consistency is a property that corresponds to asymptotic reliability (in the sense of avoiding error). Two notions of consistency are often discussed in the statistical literature — pointwise consistency and uniform consistency (see e.g. Bickel, Doksum [2001]). We first define them for tests and discuss point estimators and confidence regions later. Here is the generic setting throughout the paper[3]. $\mathbf{O}$ is the set of observed random variables, of which $O^n = \{O_1, ..., O_n\}$ denotes an i.i.d sample. $\mathcal{G}$ is the set of all possible causal graphs (defined by our background knowledge) over $\mathbf{O}$ (and possibly some other latent variables if causal sufficiency is not assumed). Given any $G \in \mathcal{G}$, $\Omega(G)$ is the set of distributions that are compatible with $G$ according to the assumptions. For example, given the Markov and Faithfulness assumptions, $\Omega(G)$ is the set of distributions that are Markov and Faithful to $G$. $\Omega_{\mathcal{G}}$ denotes the union of $\Omega(G)$'s: $\Omega_{\mathcal{G}} = \cup_{G \in \mathcal{G}} \Omega(G)$.

A (generalized) test $\phi$ is a sequence of functions $(\phi_1, \phi_2, ..., \phi_n, ...)$, where each $\phi_i$ takes data $O^i$ and returns 0, 1, or 2, representing "acceptance", "rejection" or "no conclusion", respectively. Let $\theta$ be any causal parameter of interest, which is in general a functional of the probability distribution $P$ and the causal structure $G$: $\theta = T(P, G)$. With respect to the null hypothesis $H_0 : \theta = \theta_0$ versus the alternative hypothesis $H_1 : \theta \neq \theta_0$, we define

$$\Omega_{\mathcal{G}0} = \{P : \exists G \in \mathcal{G}(P \in \Omega(G) \land T(P, G) = \theta_0)\}$$
$$\Omega_{\mathcal{G}1} = \{P : \exists G \in \mathcal{G}(P \in \Omega(G) \land T(P, G) \neq \theta_0)\}$$

Intuitively $\Omega_{\mathcal{G}i}$ is the set of distributions that are compatible with $H_i$, $i = 0, 1$. Usually $\Omega_{\mathcal{G}0}$ and $\Omega_{\mathcal{G}1}$ are not disjoint when the time order between variables is unknown. The truth value of the null hypothesis is obviously underdetermined by the distributions in the intersection of $\Omega_{\mathcal{G}0}$ and $\Omega_{\mathcal{G}1}$[4]. The inclusion of "no conclusion" in the outputs of tests respects this fact.

Let $P^n$ denote the $n$-fold product measure corresponding to $P$. The key definitions are stated below:

**Definition 1 (pointwise consistency)** *A test $\phi$ is pointwise consistent if*
*(i) for every $P \in \Omega_{\mathcal{G}0}$, $\lim_n P^n(\phi_n(O^n) = 1) = 0$ and*
*(ii) for every $P \in \Omega_{\mathcal{G}1}$, $\lim_n P^n(\phi_n(O^n) = 0) = 0$*

**Definition 2 (uniform consistency)** *A test $\phi$ is uniformly consistent if*
*(i) $\lim_n \sup_{P \in \Omega_{\mathcal{G}0}} P^n(\phi_n(O^n) = 1) = 0$ and*
*(ii) for every $\delta > 0$, $\lim_n \sup_{P \in \Omega_{\mathcal{G}1\delta}} P^n(\phi_n(O^n) = 0) = 0$*
*where*

$$\Omega_{\mathcal{G}1\delta} = \{P : \exists G \in \mathcal{G}(P \in \Omega(G) \land |T(P, G) - \theta_0| \geq \delta)\}$$

It should be clear from the definition that uniform consistency (but not pointwise consistency) allows us to simultaneously control the worst case type I error and type II error with finite sample size (given that the true parameter value is bounded away by a constant from the null value). The error bounds for a merely pointwise consistent procedure depend on the value of $\theta$, which is, unfortunately, what we want to figure out in the first place.

An obviously uniformly consistent procedure is to always return 2 in the limit. We exclude such uninformative tests by considering only non-trivial ones in the following sense[5]:

---

[2]We assume the errors are uncorrelated. Correlated Gaussian errors to a large extent can be dealt with by introducing new latent variables [Spirtes et al. 1996].

[3]Our notations and the subsequent definitions largely follow Robins et al. [2000].

[4]Even if the intersection is empty, it could also occur, in the presence of latent variables, that a distribution in $\Omega_{\mathcal{G}0}$ shares the same marginal distribution over the observed variables with a distribution in $\Omega_{\mathcal{G}1}$, in which case the hypothesis is still underdetermined by what we can observe.

[5]This definition of non-triviality is a minimal one. It could be strengthened without affecting the positive result. There is not enough room here to discuss the details, which can be found in Zhang [2002].

**Definition 3 (non-triviality)** *A test $\phi$ is non-trivial if for some $P \in \Omega_{\mathcal{G}}$,*

$$\lim_n P^n(\phi_n(O^n) = 0) = 1 \quad or \quad \lim_n P^n(\phi_n(O^n) = 1) = 1$$

## 2 UNIFORM CONSISTENCY WITH MORE FAITHFULNESS

We assume causal sufficiency in this section, and discuss the situation without the assumption in the next section.

### 2.1 A CANONICAL CASE

Consider a canonical case of (constraint-based) causal inference. In Figure 1, all variables are observed, i.e. $\mathbf{O} = \{X_1, X_2, X_3, X_4\}$, and there are no latent variables. Without further background information $\mathcal{G}$ includes all possible DAGs over $O$. In particular, $G_1, G_2 \in \mathcal{G}$. The (standardized) structural equation models associated with $G_1$ and $G_2$ are $\mathcal{M}_1$ and $\mathcal{M}_2$, respectively, as follows:

| $\mathcal{M}1$ | $\mathcal{M}2$ |
|---|---|
| $X_1 = \epsilon_1$ | $X_1 = \epsilon_1$ |
| $X_2 = \epsilon_2$ | $X_2 = \epsilon_2$ |
| $X_3 = \alpha X_1 + \beta X_2 + \epsilon_3$ | $X_3 = fX_1 + gX_2 + hX_4 + \epsilon_3$ |
| $X_4 = \gamma X_3 + \epsilon_4$ | $X_4 = mX_1 + nX_2 + \epsilon_4$ |

where all error terms are uncorrelated Gaussians with zero means. The linear coefficients in the models are naturally interpreted as the direct causal (manipulation) effects of one variable on the other. For example, in $\mathcal{M}_1$ if we manipulate $X_3$ by one unit (actually one standard deviation in the unstandardized situation) without affecting other variables unless via $X_3$, the expectation of $X_4$ will change by $\gamma$ units. (Formally, the manipulation is done by replacing the third equation in $\mathcal{M}_1$ with the equation $X_3 = c$ for some constant $c$.) So $\gamma$ quantifies the direct manipulation effect of $X_3$ on $X_4$.

Suppose we want to test $H_0 : \theta = \theta_0$ against $H_1 : \theta \neq \theta_0$, where $\theta$ is the direct manipulation effect of $X_3$ on $X_4$. $\theta$ obviously depends on the causal graphs. For example, in $G_1$ $\theta = \gamma$, while in $G_2$ $\theta = 0$. It is easy to verify that $G_1$ is the only structure that can faithfully generate the distributions such that $X_1 \perp\!\!\!\perp X_2$ and $X_4 \perp\!\!\!\perp \{X_1, X_2\}|X_3$ and no other conditional independence relations hold. In other words, $G_1$ is the only graph in the Markov equivalence class it belongs to. So, under the Faithfulness assumption, if $X_1 \perp\!\!\!\perp X_2$ and $X_4 \perp\!\!\!\perp \{X_1, X_2\}|X_3$ and no other conditional independence relations hold, $G_1$ is the true causal structure, and $\theta$ is obviously identified (with the correlation between $X_3$ and $X_4$, as $\gamma$ is). It follows that there are non-trivial pointwise consistent tests for testing $H_0 : \theta = \theta_0$ against $H_1 : \theta \neq \theta_0$ given the existence of pointwise consistent tests for correlations and partial correlations. The tests return informative answers in the large sample limit for the distributions faithful to $G_1$.

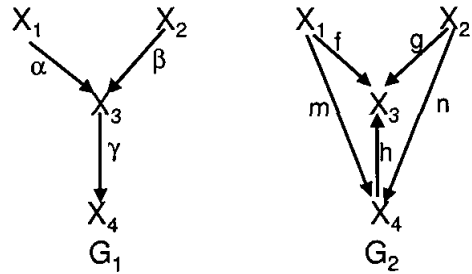

Figure 1: A Canonical Case

However, it has been shown that under the Markov and Faithfulness assumptions, there are no non-trivial uniformly consistent tests of $H_0 : \theta = \theta_0$ for $\theta_0 \neq 0$ [Robins et al. 2000, Spirtes et al. 2001], though we do have uniformly consistent tests for correlations and partial correlations[6]. The proof turns on the fact that for any distribution $P \in \Omega_{\mathcal{G}_0}$, it is possible to find a distribution $Q \in \Omega_{\mathcal{G}_1\delta}$ (for some $\delta > 0$) such that $Q$ is arbitrarily close to $P$, and vice versa.

Below we introduce two (families of) naturally strengthened versions of Faithfulness — which we call $k$-Constraint assumption and $\lambda$-Strong-Faithfulness assumption respectively — and investigate the consistency property of causal inference in the canonical case and in general under the stronger assumptions. These two families of assumptions are interesting in that they naturally generalize and on the boundary collapse into the usual Faithfulness assumption.

### 2.2 $k$-CONSTRAINT

Given a causal graph $G$, let $\Psi_G$ be the set of (standardized) linear structural coefficients for $G$ that imply covariance matrices faithful to $G$. For any (small) fixed positive constant $k$, we define $k$-Constraint as below:

**Definition 4 (k-Constraint)** *The k-Constraint is a subset $\Psi_G$ denoted by $\Psi_G^k$ such that for every $\mu \in \Psi_G$,*

$$\mu \in \Psi_G^k \iff \forall a, b \in \mathbf{O}, C \subseteq \mathbf{O}\setminus\{a,b\}(|\rho_{ab.C}| \geq k|\mu_{ab}|)$$

*where $\rho_{ab.C}$ is the partial correlation between $a$ and $b$*

---

[6] For multivariate Gaussian distributions, the test of correlation based on Fisher's Z-transformation, for example, is uniformly consistent. There is also a very nice relationship between the inference of partial correlation and the inference of correlation. See, e.g. Anderson [1958] p78, 85.

given $C$ (or correlation when $C = \emptyset$) in the distribution generated by $(\mu, G)$, and $\mu_{ab}$ is the coefficient associated with the arrow, if any, between $a$ and $b$.

The $k$-**Constraint assumption** says that for every $G \in \mathcal{G}$, if the true structure is $G$, then the true structural parameter is in $\Psi_G^k$. It is easy to see that when $k = 0$, the $k$-Constraint assumption adds nothing more to the Faithfulness assumption.

Under the Gaussian parameterizations, the usual Faithfulness assumption entails that for any two variables $a$ and $b$, if $\rho_{ab.C} = 0$ for some $C \subseteq \mathbf{O}\setminus\{a, b\}$, there cannot be an arrow between $a$ and $b$, and hence there is no direct causal effect between $a$ and $b$. In short, vanishing (partial) correlations indicate no (direct) effects. The $k$-Constraint assumption assumes furthermore that *small* (partial) correlations indicate *small* (direct) causal effects. In this sense it is a natural generalization of the Faithfulness assumption. In practice, it is not uncommon among social scientists to both interpret the regression coefficients causally and delete insignificant variables based on, say, the p-values of $t$ tests. A charitable interpretation is that they implicitly adopt (something like) the $k$-Constraint assumption: small correlation means small effect.

Under the Markov and $k$-Constraint assumptions (no matter how small $k$ is), we can construct non-trivial uniformly consistent procedures to test $H_0 : \theta = \theta_0$ in the canonical case by modifying the typical constraint-based algorithms (such as the PC algorithm) slightly. We will describe the modification after we introduce the $\lambda$-Strong-Faithfulness, where more intuition can be gained[7]. To see why the existing algorithms in the literature have to be modified to be uniformly consistent, we present a negative result here that may give a hint.

Suppose, in the canonical case, the background knowledge is sufficient for us to conclude that the true causal graph is either $G_1$ or $G_2$, i.e. $\mathcal{G} = \{G_1, G_2\}$. Under this circumstance, $\Omega_{\mathcal{G}_0} \cap \Omega_{\mathcal{G}_1} = \emptyset$, that is, there is no issue of underdetermination. Clearly a test does not need the answer of "no conclusion" in order to be (pointwise) consistent. However, there are no uniformly consistent tests that do not return "no conclusion" under the Markov and $k$-Constraint assumptions, which is a direct consequence of the following theorem.

**Theorem 1** *Given the Markov assumption and the $k$-Constraint assumption, for any $\theta_0 \neq 0$, if $G_1, G_2 \in \mathcal{G}$, there is no uniformly consistent test $\phi$ for $H_0 : \theta = \theta_0$ versus $H_1 : \theta \neq \theta_0$ such that*

$$\lim_{n \to \infty} \sup_{P \in \Omega_{G_1 0}} P^n(\phi_n(O^n) = 1 \lor \phi_n(O^n) = 2) = 0$$

---

[7] The formal proof can be found in Zhang [2002], which we leave out for lack of space.

*where $\Omega_{G_1 0}$ is the set of all legitimate Gaussian distributions generated by $G_1$ with $\gamma = \theta_0$*

This result applies to any test procedures, including, for example, tests based on the various model selection scores. Intuitively, it is very difficult to recover $G_1$ when it is the true graph but the arrow between $X_1$ and $X_3$ or the one between $X_2$ and $X_3$ is very weak, because it is then hard to detect the correlation between $X_1$ and $X_3$ or that between $X_2$ and $X_3$. The $k$-Constraint assumption does not bound $\alpha$ and $\beta$ away from zero. The fact that they can become arbitrarily small is responsible for the lack of uniform consistency of the typical algorithms in the literature that only test whether some correlations are vanishing. We have to modify those algorithms to control the nuisance parameters $\alpha, \beta$ somehow in order to obtain uniform consistency, as described later.

### 2.3 $\lambda$-STRONG-FAITHFULNESS

A perhaps more direct strengthening of Faithfulness is the following:

**Definition 5 ($\lambda$-Strong-Faithfulness)** *A Gaussian distribution $P$ is said to be $\lambda$-Strong-Faithful to a DAG $G$ with observed variables $O$ if for any $a, b \in O$ and $C \subseteq O\setminus\{a, b\}$,*

*$a$ is d-connected to $b$ given $C$ in $G$ $\iff$ $|\rho_{ab.C}| > \lambda$*

*where $\lambda \in (0, 1)$ is a fixed (small) constant.*

Obviously 0-Strong-Faithfulness is just the usual Faithfulness. The $\lambda$-**Strong-Faithfulness assumption** says that the Gaussian distributions generated by a causal graph are $\lambda$-Strong-Faithful to the graph. Intuitively the difference between $\lambda$-Strong-Faithfulness and $k$-Constraint is that $\lambda$-Strong-Faithfulness further rules out the possibility of weak arrows in the graph. For example, it entails that $|\alpha| > \lambda, |\beta| > \lambda$ in $G_1$. Under the $\lambda$-Strong-Faithfulness assumption (no matter how small $\lambda$ is), the inference of causal structure in general can be uniformly consistent.

**Theorem 2** *Let $\mathcal{M}$ be an arbitrary Markov equivalence class. Consider the null hypothesis $H_0$: data $O^n$ are generated from a structure in $\mathcal{M}$. Given the Markov and $\lambda$-strong-faithfulness assumptions, there exists a test $\phi$ of the null hypothesis such that $\phi$ only returns 0 or 1 and*

$$\lim_n \sup_{P \in \Gamma_0} P^n(\phi_n(O^n) = 1) = 0$$

$$\text{and} \quad \lim_n \sup_{P \in \Gamma_1} P^n(\phi_n(O^n) = 0) = 0$$

*where*

$$\Gamma_0 = \bigcup_{G \in \mathcal{M}} \Omega(G) \qquad \Gamma_1 = \bigcup_{G \notin \mathcal{M}} \Omega(G)$$

The test constructed in the proof (in the appendix) is a combination of a series of tests of vanishing partial correlations, which is exactly what the constraint-based algorithms suggest. In other words, those algorithms are uniformly consistent without further modification under the $\lambda$-Strong-Faithfulness assumption. Furthermore, it follows from Theorem 2 that as long as a causal parameter can be identified in some Markov equivalence class, there exist non-trivial uniformly consistent tests for that parameter provided that there are uniformly consistent tests for the corresponding statistical quantity.

In view of Theorem 1, it is easy to show that Theorem 2 cannot be true if the $\lambda$-Strong-Faithfulness assumption is replaced by the $k$-Constraint assumption. As noted earlier, unlike the $\lambda$-Strong-Faithfulness assumption, the $k$-Constraint assumption allows the possibility of arbitrarily weak arrows, which act as nuisance parameters when they are not of direct interest. In the canonical case, for instance, $\alpha$ and $\beta$ are nuisance parameters, which we have to control somehow in order to get uniformly consistent tests. This requires modifying the existing algorithms that only involve testing vanishing partial correlations. One way to modify the algorithm is to further test the null hypotheses $|\rho_{X_1 X_3}| \geq \alpha_0$ and $|\rho_{X_2 X_3}| \geq \beta_0$ for fixed $\alpha_0, \beta_0$ and return "no conclusion" if any of the tests rejects the null hypotheses (see Zhang [2002] for details). Technically this modification amounts to building some $\lambda$-Strong-Faithfulness into the algorithm, as the modified algorithm refuses to give informative answers when the correlations are below some threshold. The modified algorithms are clearly less informative as they return "no conclusions" under more circumstances, which is what one has to pay for adopting a less stringent (but more plausible) assumption.

## 2.4 ESTIMATORS AND CONFIDENCE REGIONS

There is nothing special about tests. The foregoing discussions on consistency can also be formulated in terms of point estimators and confidence regions[8]. Since a causal parameter is only sometimes identifiable, we need also to generalize the notions of point estimators and confidence regions, just as we include an uninformative answer in the outputs of tests. Robins et al. [2000] defined a generalized estimator $\hat{\theta}$ of $\theta$ as a sequence $(\hat{\theta}_1, ..., \hat{\theta}_n, ...)$, where each $\hat{\theta}_i$ is a function of $O^i$ that returns a non-empty subset of the parameter space $\Theta$. A non-singleton set (e.g. $\Theta$) indicates that no point estimation can be made, though non-trivial bounds may be given in some settings[9]. Let $\Omega\mathcal{G} = \{(P,G) : G \in \mathcal{G}, P \in \Omega(G)\}$. $\hat{\theta}$ is **pointwise consistent** if for every $(P,G) \in \Omega\mathcal{G}$, $\hat{\theta}$ converges in probability to $\theta = T(P,G)$, namely, for every $\epsilon > 0$,

$$\lim_{n\to\infty} P^n(d[\hat{\theta}_n(O^n), T(P,G)] > \epsilon) = 0$$

$\hat{\theta}$ is said to be **uniformly consistent** if for every $\epsilon > 0$,

$$\lim_{n\to\infty} \sup_{(P,G)\in\Omega\mathcal{G}} P^n(d[\hat{\theta}_n(O^n), T(P,G)] > \epsilon) = 0$$

where the distance $d$ between a set $S$ and a real number $r$ is defined as the shortest Euclidean distance between $r$ and the elements in $S$: $d[S,r] = \inf_{s\in S} |s-r|$. Finally $\hat{\theta}$ is **non-trivial** if for some $P \in \Omega_\mathcal{G}$

$$\lim_{n\to\infty} P^n(\hat{\theta}_n(O^n) \text{ is a singleton}) = 1$$

In the constraint-based causal inference, a typical estimator of a causal parameter first pins down a Markov equivalence class via a series of tests of independence and conditional independence relations, and then estimates the parameter if it is identifiable in the resulting equivalence class or returns the whole sample space otherwise. In view of Theorem 2, it is not hard to see that this estimator is uniformly consistent under the Markov and $\lambda$-strong-faithfulness assumptions, provided that the estimator of the statistical quantity, with which the causal parameter is identified, is uniformly consistent.

We define the generalized confidence regions in a way that maintains the well known duality between tests and confidence regions. Let $\mathbf{R}_\alpha$ denote a sequence: $(\mathbf{R}_{\alpha,1}, ..., \mathbf{R}_{\alpha,n}...)$, where each $\mathbf{R}_{\alpha,i}$ is a function of $O^i$ which returns a triple partition $(S^0_{\alpha,i}, S^1_{\alpha,i}, S^2_{\alpha,i})$ of the parameter space $\Theta$. $\mathbf{R}_\alpha$ is called a (generalized) $1-\alpha$ **confidence region** of $\theta = T(P,G)$ if

$$\liminf_n \inf_{(P,G)\in\Omega\mathcal{G}} P^n(T(P,G) \notin S^1_{\alpha,n}) \geq 1 - \alpha$$

The definition is obviously given with test inversion in mind. We can easily invert a test (actually a family of tests) into a confidence region thus defined: $S^0_{\alpha,n}$ contains the values of the parameter that are accepted, $S^1_{\alpha,n}$ contains the values rejected and $S^2_{\alpha,n}$ contains the rest, for which "no conclusion" are returned.

$\mathbf{R}_\alpha$ is said to be **pointwise consistent** if for every $(P,G), (Q,H) \in \Omega\mathcal{G}$ such that $T(P,G) \neq T(Q,H)$,

$$\lim_{n\to\infty} P^n(T(Q,H) \in S^0_{\alpha,n}) = 0$$

it is **uniformly consistent** if for every $\delta > 0$

$$\lim_{n\to\infty} \sup_{|T(Q,H)-T(P,G)|>\delta} P^n(T(Q,H) \in S^0_{\alpha,n}) = 0$$

---
[8] Many details are left out in this section, which can be found in Zhang [2002]

[9] We thank an anonymous referee for pointing out this.

it is **non-trivial** if at least for some $(P, G) \in \Omega \mathcal{G}$,

$$\lim_{n \to \infty} P^n(S^2_{\alpha,n} \neq \Theta) = 1$$

It is not hard to verify the usual duality between tests and confidence regions in this generalized setting. In particular, a family of uniformly consistent tests can be inverted into a uniformly consistent $1 - \alpha$ confidence region for any $\alpha \in (0, 1)$, at least in principle. Hence under the strong-faithfulness assumptions, we can hope for uniformly consistent confidence regions of causal parameters.

It may appear difficult to interpret the generalized confidence regions, but in most cases, fortunately, the test inversion will lead to a confidence region such that either $S^2_{\alpha,n} = \emptyset$ (when $\theta$ is identifiable), or $S^0_{\alpha,n} = S^1_{\alpha,n} = \emptyset$ (when $\theta$ is not identifiable). For example, in the canonical case, either the causal structure suggested by data is $G_1$ in which case we can calculate an informative confidence interval, or the structure indicated by data is not $G_1$ in which case the confidence interval is the uninformative one, the whole set of legitimate values. Either way the resulting confidence interval looks just like the ordinary confidence interval. The only cases where the generalized confidence regions are non-standard are where some value of the parameter (usually 0) is of special status.

## 3 DISCUSSION

### 3.1 WITHOUT CAUSAL SUFFICIENCY

It should be clear that the proof of Theorem 2 does not depend on the assumption of causal sufficiency at all. So, even in the presence of latent confounders, the typical causal inference algorithms, such as FCI in Spirtes et al. [1993], are uniformly consistent under the Markov and $\lambda$-Strong-Faithfulness assumptions. Under the $k$-Constraint assumption as currently defined, however, causal sufficiency is in general necessary to guarantee the possibility of uniformly consistent causal inference[10]. It is possible nonetheless to define the $k$-Constraint with respect to the parameterization of Maximal Ancestral Graphs (MAGs) [Richardson, Spirtes 2000] so that uniform consistency can be established without causal sufficiency. A problem with such a definition is that the nice intuitive explanation of $k$-Constraint — small (partial) correlation indicates small (direct) effect — is no longer available, as the parameters in MAGs do not always correspond to direct causal effects. We have not yet figured out a nice intuition behind the $k$-Constraint assumption defined over MAGs, which is certainly an interesting question for future work.

### 3.2 SOME REFLECTION ON STRONG FAITHFULNESS

The two strong-faithfulness assumptions laid out in this paper are both indexed by a positive constant. No matter how small the constant is, the assumptions entail the possibility of uniformly consistent causal inference. Both assumptions collapse into the faithfulness assumption on the boundary: 0-Constraint is a vacuous constraint and 0-Strong-Faithfulness is essentially the same as faithfulness.

The perhaps most powerful and frequently used defense for the Faithfulness assumption is that under the Gaussian or the multinomial parameterization, given a causal structure $G$, the set of parameters that lead to distributions unfaithful to $G$ has zero Lebesgue measure (Spirtes et al. 1993, Meek 1995). A nice consequence of this fact is that for any causal structure $G$, any prior that is absolutely continuous with respect to Lebesgue measure will assign 0 probability to the unfaithful distributions. It is not necessarily the case that the Lebesgue measure of the set of parameters ruled out by, for example, the $\lambda$-Strong-Faithfulness assumption can be made arbitrarily small (by decreasing $\lambda$) unless the parameter space is bounded. But it is true that given a causal structure $G$ and a prior absolutely continuous with respect to Lebesgue measure, for any $\epsilon > 0$, there exists a $\lambda$ such that the prior probability assigned to the set of distributions that are not $\lambda$-strong-faithful to $G$ is less than $\epsilon$. This readily follows from the continuity of the probability measure.

An implication of the "measure 0" result is the existence of faithful multivariate Gaussian distributions (and multinomial distributions) for every causal structure, which is also cited fairly often in the literature. It is certainly not the case that for every $\lambda \in (0, 1)$ and every causal structure $G$, there exists a distribution $\lambda$-Strong-Faithful to $G$.[11] For example, if $A$, $B$, $C$ are three independent Gaussian parents of $D$, it is impossible that the correlations between $D$ and each of the parents are all greater than $\sqrt{3}/3$. On the other hand, it is trivial to see that for any causal structure $G$, there exists a multivariate Gaussian distribution $\lambda$-Strong-Faithful to $G$ for some $\lambda$. More interestingly, it can be shown that given a fixed set of observed variables $O$, we can find a small $\lambda$ such that for every causal structure $G$ with $O$ as the observed variables, there exists a multivariate Gaussian distribution $\lambda$-Strong-Faithful to $G$. The magnitude of $\lambda$ depends on the number of variables in $O$.

---

[10]In the canonical case, for example, if we allow the possibility that there might be a latent confounder between $X_1$ and $X_4$, it can be shown fairly easily that there is no non-trivial uniformly consistent test even under the $k$-Constraint assumption.

[11]It is not yet clear whether for every $k \in (0, 1)$ and every causal structure $G$, there exists a Gaussian distribution that satisfies the $k$-constraint with $G$.

Another popular interpretation of the Faithfulness assumption appeals to the notion of "stability". [Pearl 2000] The faithful distributions are stable in the sense that the independence and conditional independence relations associated with the distributions cannot be destroyed by small variations of parameters. Similarly, a faithful but close to unfaithful distribution may be said to be unstable in the sense that some dependence relations may be destroyed by a slight change in parameterization. In this sense, the $\lambda$ in the $\lambda$-Strong-Faithfulness serves as a rough index of stability.

It is not the main purpose of this reflection to argue for the plausibility of the strong-faithfulness assumptions. Rather the discussion is to illustrate the close relation between the usual faithfulness condition and the stronger faithfulness conditions laid out in the paper. Clearly in several important respects, the faithfulness assumption is just a limiting case of the $\lambda$-Strong-Faithfulness assumption (or the $k$-Constraint assumption). This suggests that the stronger assumptions are not only sufficient but also close to necessary to entail the existence of uniformly consistent causal inference procedures without substantial background knowledge.

### Acknowledgement

We thank Clark Glymour and Tianjiao Chu for helpful discussions, and the referees for valuable comments. The research was supported by the following NASA grants: NCC-2-1377, NCC-2-1295, and NCC-1-1227.

### Appendix

**Proof of Theorem 1.**

**Lemma** Consider the standardized structural equation models $M_1$ and $M_2$ introduced in section 2.1. For every $0 < k < 1$, $\theta_0 \neq 0$ and every $\epsilon > 0$, there are $\mu_1 = (\alpha, \beta, \gamma = \theta_0)$ and $\mu_2 = (f, g, h, m, n)$ such that $\mu_i \in \Psi_{G_i}^k$ and $KL(P_1, P_2) < \epsilon$, where $P_i$ is the distribution generated by $(\mu_i, G_i)$, and $KL(P_1, P_2)$ is the Kullback-Leibler divergence between $P_1$ and $P_2$.

**Proof** The correlation matrix generated by $M_1$ is

$$\Sigma_1 = \begin{pmatrix} 1 & 0 & \alpha & \alpha\gamma \\ & 1 & \beta & \beta\gamma \\ & & 1 & \gamma \\ & & & 1 \end{pmatrix}$$

It is easy to verify that all legitimate parameters of $M_1$ are in $\Psi_{G_1}^k$, i.e. no (more) constraints are placed on the parameters in $M_1$ by the $k$-constraint. The correlation matrix generated by $M_2$ is

$$\Sigma_2 = \begin{pmatrix} 1 & 0 & f+mh & m \\ & 1 & g+nh & n \\ & & 1 & fm+gn+h \\ & & & 1 \end{pmatrix}$$

The $k$-constraint assumption puts the following constraints on $\mu_2$:

$$k|m| \leq \left| \frac{m - (f+mh)(fm+gn+h)}{\sqrt{(1-(f+mh)^2)(1-(fm+gn+h)^2)}} \right|$$

$$
\begin{aligned}
&= |\rho_{X_1 X_4 . X_3}| \\
k|n| &\leq \left| \frac{n - (g+nh)(fm+gn+h)}{\sqrt{(1-(g+nh)^2)(1-(fm+gn+h)^2)}} \right| \\
&= |\rho_{X_2 X_4 . X_3}| \\
k|f| &\leq |f + mh| = |\rho_{X_1 X_3}| \\
k|g| &\leq |g + nh| = |\rho_{X_2 X_3}| \\
k|h| &\leq |fm + gn + h| = |\rho_{X_3 X_4}|
\end{aligned}
$$

Given any $\epsilon > 0$, there exists $\delta_{\alpha,\beta} > 0$ ($\delta_{\alpha,\beta}$ depends on $\alpha$ and $\beta$) such that if

$$
\begin{aligned}
f + mh &= \alpha & (1) \\
g + nh &= \beta & (2) \\
fm + gn + h &= \gamma = \theta_0 & (3) \\
|m - \alpha\theta_0| &< \delta_{\alpha,\beta} & (4) \\
|n - \beta\theta_0| &< \delta_{\alpha,\beta} & (5)
\end{aligned}
$$

(i.e. the correlation matrices are close enough) then $KL(P_1, P_2) < \epsilon$. Solve (1),(2),(3) for $f, g, h$, we get

$$
\begin{aligned}
f &= \frac{(1-n^2)\alpha + mn\beta - m\theta_0}{1 - m^2 - n^2} \\
g &= \frac{(1-m^2)\beta + mn\alpha - n\theta_0}{1 - m^2 - n^2} \\
h &= \frac{\theta_0 - m\alpha - n\beta}{1 - m^2 - n^2}
\end{aligned}
$$

It is not hard to check that we can choose appropriate (small) $m, n, \alpha, \beta$ to satisfy (4), (5) and all the constraints. **Q.E.D**

**Proof of Theorem 1** For the sake of contradiction, suppose there is such a test $\phi$. Choose $\delta < \theta_0$ so that $\Omega_{\mathcal{G}_1\delta}$ includes all the distributions in $\Omega_{\mathcal{G}_2}$. Now given any $\epsilon > 0$, by the above Lemma, there are $P_1 \in \Omega_{\mathcal{G}_1}$ and $P_2 \in \Omega_{\mathcal{G}_2}$ such that $KL(P_1^n, P_2^n) < 4\epsilon^2$. Hence, by the relationship between KL divergence and total variation distance,

$$\sup_E |P_1^n(E) - P_2^n(E)| \leq 1/2\sqrt{KL(P_1^n, P_2^n)} = \epsilon$$

The supremum is over all the events in the sample space. Therefore,

$$
\begin{aligned}
\sup_{R \in \Omega_{\mathcal{G}_1\delta}} & R^n(\phi_n(O^n) = 0) \\
&\geq P_2^n(\phi_n(O^n) = 0) \\
&\geq P_1^n(\phi_n(O^n) = 0) - \epsilon \\
&= 1 - P_1^n(\phi_n(O^n) = 1 \vee \phi_n(O^n) = 2) \\
&\geq 1 - \sup_{P \in \Omega_{\mathcal{G}_1 0}} P^n(\phi_n(O^n) = 1 \vee \phi_n(O^n) = 2) - \epsilon \\
&\to 1 - \epsilon
\end{aligned}
$$

Note that the selection of $\epsilon$ is arbitrary, which implies

$$\lim_{n \to \infty} \sup_{R \in \Omega_{\mathcal{G}_1\delta}} R^n(\phi_n(O^n) = 0) = 1$$

This is a contradiction as $\phi$ is assumed to be uniformly consistent. **Q.E.D**

**Proof of Theorem 2.**

$\mathcal{M}$ entails a unique set of vanishing partial correlations, and there are altogether a finite number of partial correlations to consider among the observed variables. We construct a test $\phi$ as below:

$$\phi_n(O^n) = \begin{cases} 0 & \text{if } T_n^1(O^n) = \ldots = T_n^l(O^n) = 0 \\ & \text{and } T_n^{l+1}(O^n) = \ldots = T_n^m(O^n) = 1 \\ 1 & \text{otherwise} \end{cases}$$

where $T^1, \ldots, T^l$ are the uniformly consistent tests of vanishing partial correlations for the ones entailed to be zero by $\mathcal{M}$, and $T^{l+1}, \ldots, T^m$ are the tests of vanishing partial correlations for the rest. Generically the null hypothesis of $T^i$ is $\theta_i = 0$, where $\theta_i$ is a partial correlation. We show that $\phi$ satisfies the requirement. Obviously it only returns 0 or 1. Let

$$
\begin{aligned}
p_i^n &= \sup_{P \in \mathcal{H}_{1\lambda}^i} P^n(T^i(O^n) = 0), \quad i = 1, \ldots, l \\
p_j^n &= \sup_{P \in \mathcal{H}_0^j} P^n(T^j(O^n) = 1), \quad j = l+1, \ldots, m
\end{aligned}
$$

where $\mathcal{H}_{1\lambda}^i$ is the set of distributions compatible with $|\theta_i - 0| > \lambda$, $i = 1, \ldots, l$, and $\mathcal{H}_0^j$ is the set of distributions compatible with the null hypothesis $\theta_j = 0$, $j = l+1, \ldots, m$. Since $T^1, \ldots, T^m$ are uniformly consistent, we have

$$\lim_{n \to \infty} p_i^n = 0, i = 1, \ldots, m.$$

Under the $\lambda$-strong-faithfulness assumption, it is not hard to see that,

$$P^n(\phi_n(O^n) = 0) \leq \max\{p_i^n : i = 1, \ldots, m\}$$

Therefore

$$\sup_{P \in \Gamma_1} P^n(\phi_n(O^n) = 0) \leq \max\{p_i^n : i = 1, \ldots, m\} \to 0$$

To show that

$$\lim_n \sup_{P \in \Gamma_0} P^n(\phi_n(O^n) = 1) = 0$$

it suffices to replace max in the foregoing argument with summation, and change $p_i^n$ to $\sup_{P \in \mathcal{H}_0^i} P^n(T^i(O^n) = 1), i = 1, \ldots, l$ and $p_j^n$ to $\sup_{P \in \mathcal{H}_{1\lambda}^j} P^n(T^j(O^n) = 0), j = l+1, \ldots, m$. **Q.E.D**